\documentclass[10pt,twocolumn,letterpaper]{article}

\usepackage{cvpr}
\usepackage{times}
\usepackage{epsfig}
\usepackage{graphicx}
\usepackage{amsmath}
\usepackage{amssymb}
\usepackage{subfig}
\usepackage{color}

\usepackage{enumitem}

\usepackage[pagebackref=true,breaklinks=true,letterpaper=true,colorlinks,bookmarks=false]{hyperref}

\cvprfinalcopy 

\def\cvprPaperID{1473} 

\ifcvprfinal\fi
\begin{document}

\title{Discovering Human Interactions in Videos with Limited Data Labeling}

\author{Mehran Khodabandeh, Arash Vahdat, Guang-Tong  Zhou, Hossein  Hajimirsadeghi\\
Simon Fraser University,\\
Burnaby, BC, Canada\\
{\tt\small  \{mkhodaba, avahdat, gza11, hosseinh \}@cs.sfu.ca}
\and
Mehrsan Javan Roshtkhari \\
SPORTLOGiQ\\
Montreal, QC, Canada\\
{\tt\small mehrsan@sportlogiq.com}
\and
Greg Mori \\
Simon Fraser University,\\
Burnaby, BC, Canada\\
{\tt\small mori@cs.sfu.ca}
\and
Stephen  Se \\
MDA Corporation, \\
Richmond, BC, Canada\\
{\tt\small SSE@mdacorporation.com}
}

\maketitle

{\color{red} }

\begin{abstract}

  We present a novel approach for discovering human interactions in
  videos. Activity understanding techniques usually require a large
  number of labeled examples, which are not available in many
  practical cases. Here, we focus on recovering semantically
  meaningful clusters of human-human and human-object interaction in
  an unsupervised fashion.  A new iterative solution is
  introduced based on Maximum Margin Clustering (MMC), which also accepts
  user feedback to refine clusters. This is achieved by formulating
  the whole process as a unified constrained latent max-margin
  clustering problem. Extensive experiments have been carried out over
  three challenging datasets, Collective Activity, VIRAT, and
  UT-interaction. Empirical results demonstrate that the
  proposed algorithm can efficiently discover perfect semantic clusters of human
  interactions with only a small amount of labeling effort.

\end{abstract}

\section{Introduction}

\begin{figure}
\includegraphics[width=0.5\textwidth]{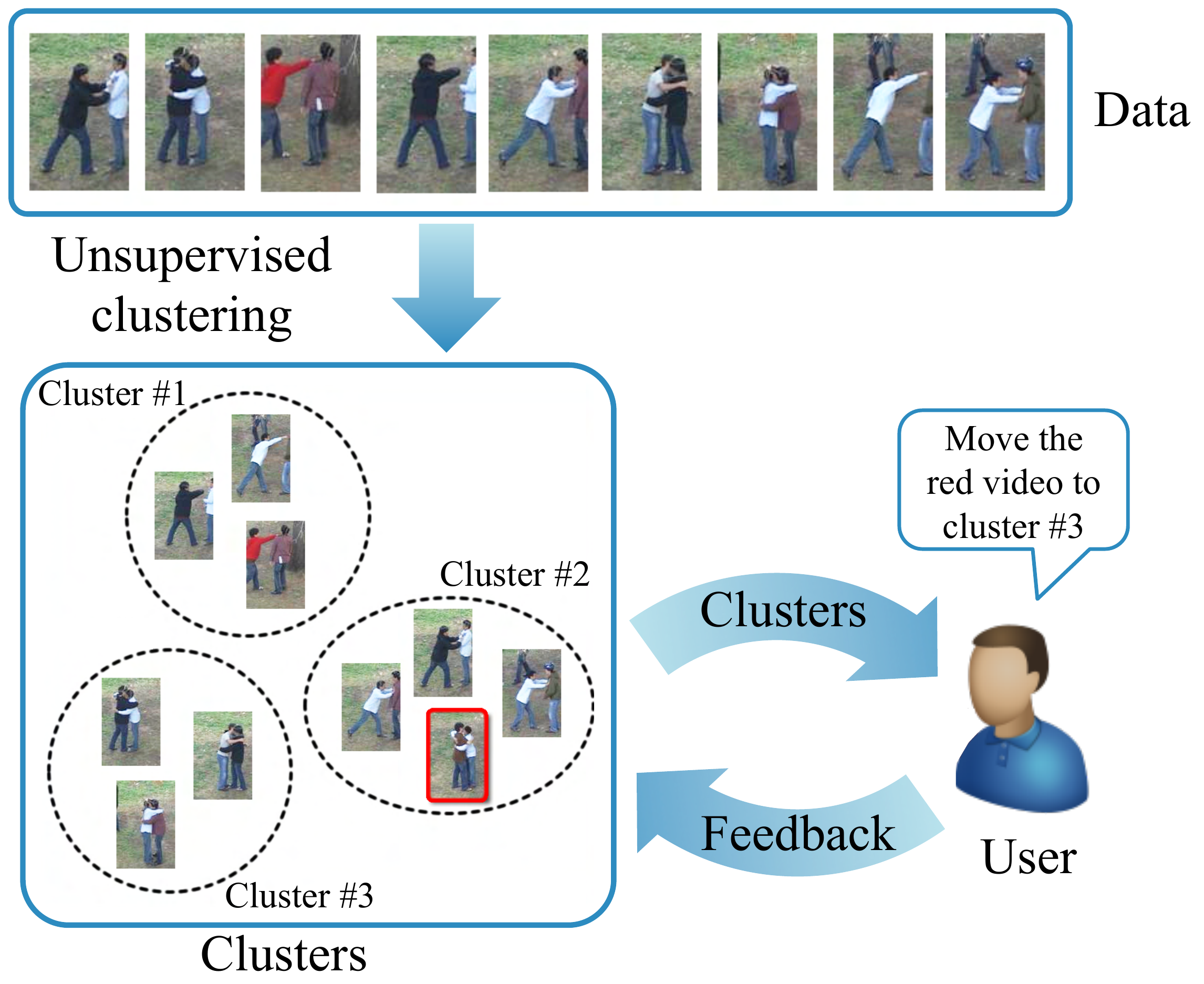}
\caption{Given a set of videos, we aim to extract clusters
  corresponding to different types of human interactions.  The top
  shows a set of sequences of human-human interactions. In the first step,
  an unsupervised clustering approach is used to create an initial set of clusters.
  Next, user feedback is obtained iteratively to refine the clusters.}
\label{fig:fig1}
\end{figure}

Automated analysis of videos of human activity can take many forms --
answering questions about the presence of specific types of activities
through to the discovery of what has happened in a scene.  In this
paper we focus on the latter and present an algorithm to label human
interactions\footnote{The term ``interaction'' 
refers to any kind of interaction between humans, and humans and objects that are present in
the scene, such as vehicles, rather than activities which 
are performed by a single subject.  } in videos.  
The algorithm works in a
clustering paradigm, starting with an unsupervised step that forms
groups of similar human interactions.  These clusters are refined
based on user feedback, and the process is iterated, as shown in Fig.~\ref{fig:fig1}.

Different strategies can be followed in order to label
how people are interacting in a set of input videos.
Brute-force labeling approaches involving manual
labour are costly, since input videos often cover a long
period of time. Hence, a common approach is to use
supervised learning and focus on detecting a set of pre-specified
activities of interest
(e.g.~\cite{MedioniCBHN01,ZhuNR13,RyooA11}). For instance, an
algorithm can be pre-trained to detect instances of people getting
into vehicles, and then find all instances of that specific
event. To obtain high accuracy, those approaches often require lots of labeled training data,
which is not easy to obtain in many cases. 

Extensions based on active learning can be used to build up a
collection of labeled data, while being efficient with human labeling
effort (e.g.~\cite{YanYH03,BandlaG13,VondrickR11}).  Impressive
results have been obtained by these supervised methods, however, these
remain limited to pre-specified categories of events.

On the other hand, unsupervised analysis techniques aim to obtain
clusters of human activities or perform novelty/outlier detection
to find rare events.  This paradigm is attractive since it requires neither
a priori specification of events nor human labeling effort.
Effective methods in this vein have been developed previously
(e.g.~\cite{HospedalesGX09,ZhongSV04,WangMG09,MahadevanLBV10,MehranOS09,RoshtkhariL13,morris2011trajectory}).
 In general, those methods focus on either creating one (or a few) big
 clusters or a large number of clusters of common activities. In the 
 former, those clusters do not necessarily represent activities of 
 the same labels and in the latter, there are 
 many clusters that are representing the same type of activity.
Our work follows in this line, but is focused on discovering and
labeling common human interactions, utilizing a clustering approach 
to create meaningful activity/interaction groups and accepting user 
feedback to improve accuracy.

In this paper we propose a novel algorithm for discovering
human interactions in video sequences.  The algorithm performs
iterated clustering and incorporation of user feedback.  The
contributions include a principled formulation of this process as a
constrained latent max-margin clustering problem.  We demonstrate that
this algorithm can be very effective, obtaining state of the art
clustering results from no labeled data, and obtaining perfect
clustering after a small amount of user feedback.

\section{Previous Work}
\label{sec:pw}

Human activity and interaction understanding is an active research area. Recent
surveys such as Poppe~\cite{Poppe10} and Weinland et
al.~\cite{WeinlandRB10} provide an overview of the literature. We emphasize 
that the objective of this work is to describe human interactions rather than individual
 activities performed by a single subject. 


\subsection{Supervised Activity Recognition}

There is an extensive literature on recognizing interactions or
analyzing the behaviours of groups of people. Much of this work
involves supervised learning, either in the form of specific classes
of interactions to detect or templates/rules for detecting
interactions of interest. Initial work in this vein includes Medioni
et al.~\cite{MedioniCBHN01}, who analyzed vehicle trajectories, for
instance detecting vehicles approaching or avoiding road checkpoints.
Intille and Bobick~\cite{IntilleB01} developed probabilistic graphical
models for interpreting football plays based on player trajectories.

Ryoo and Aggarwal~\cite{RyooA11} use stochastic grammars to compose
sub-events and the actions of individuals into larger events.
Zhu et al.~\cite{ZhuNR13} develop a method for detecting specified
human-vehicle interactions based on spatio-temporal contextual models.
Amer et al.~\cite{AmerXZTZ12} model activities at varying levels of
detail, formulating AND-OR graph representations that permit efficient inference.
Choi and Savarese~\cite{ChoiS12} develop a unified framework for
tracking and inferring the actions/activities of a group of people.
Khamis et al.~\cite{khamis-eccv2012} include temporal analysis of the
actions of individuals and develop efficient inference techniques for
analyzing collective activities.
Lan et al.~\cite{LanWYRM12} model interactions between individuals in
a scene and their relations to an over-arching scene-level activity label.
Patron-Perez et al.~\cite{Patron-PerezMZR10} detect human-human
interactions in television shows using a structural SVM approach.  Our
work builds on these methods for analyzing interactions, but aims for
unsupervised learning or discovery of interactions rather than the
supervised approach common to these methods.

Active learning approaches involve human labeling, with a learning
algorithm typically presenting the most uncertain or most helpful
unlabeled data to a user to acquire additional labels.  This type of
learning has been deployed in the object/action recognition literature, e.g.\
\cite{BandlaG13,KovashkaVG11,DuanPCG12}.  Our approach shares similarities, though
is focused on interaction {\em discovery}, within a clustering
paradigm rather than supervised recognition approach.

\begin{figure*}
\centering
\subfloat[][Initial clusters (completely unsupervised)]
{\includegraphics[scale=0.2]{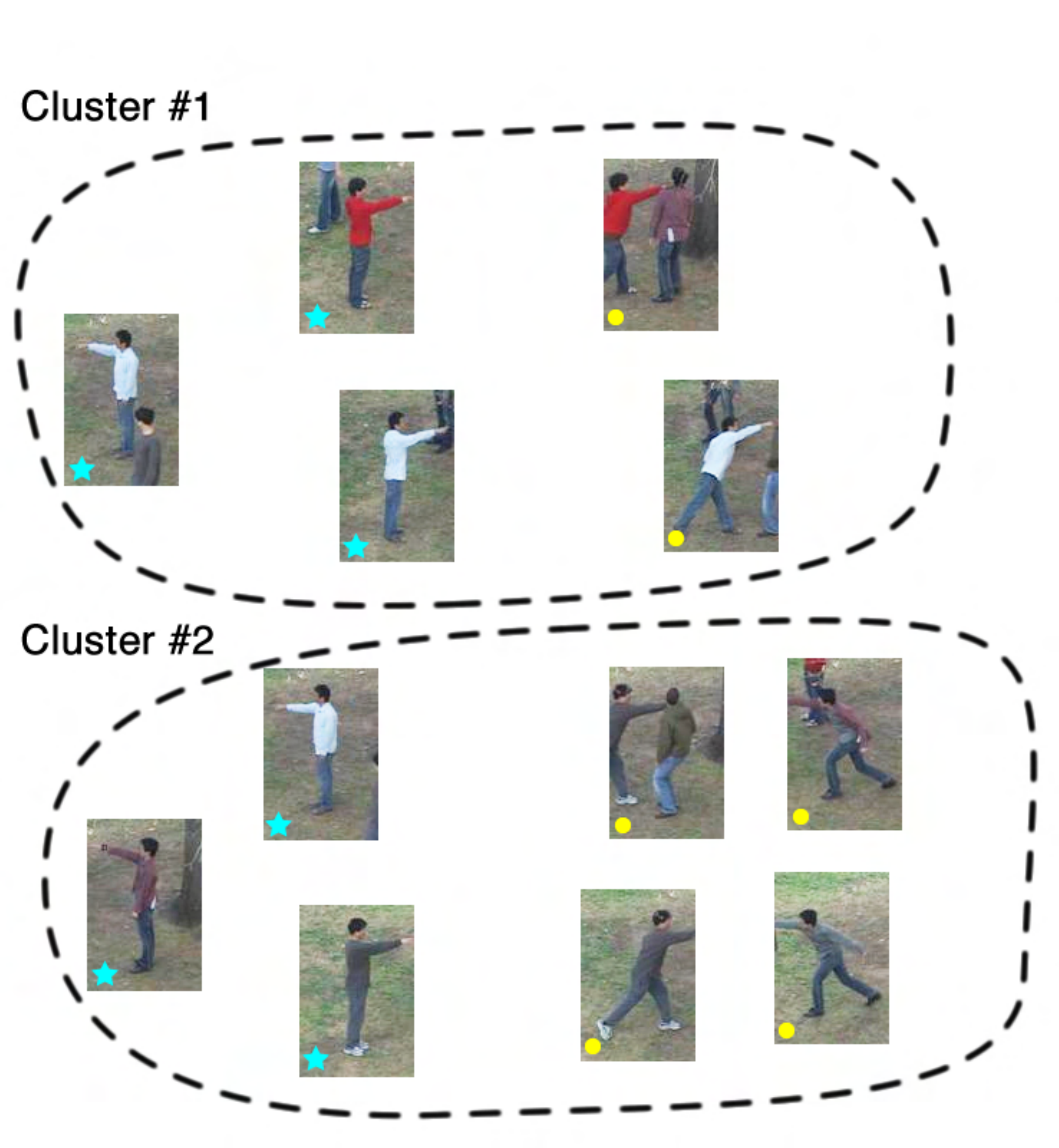}\label{fig:iter0}}
\quad
\subfloat[][\textit{Must-link} and \textit{cannot-link} samples]
{\includegraphics[scale=0.2]{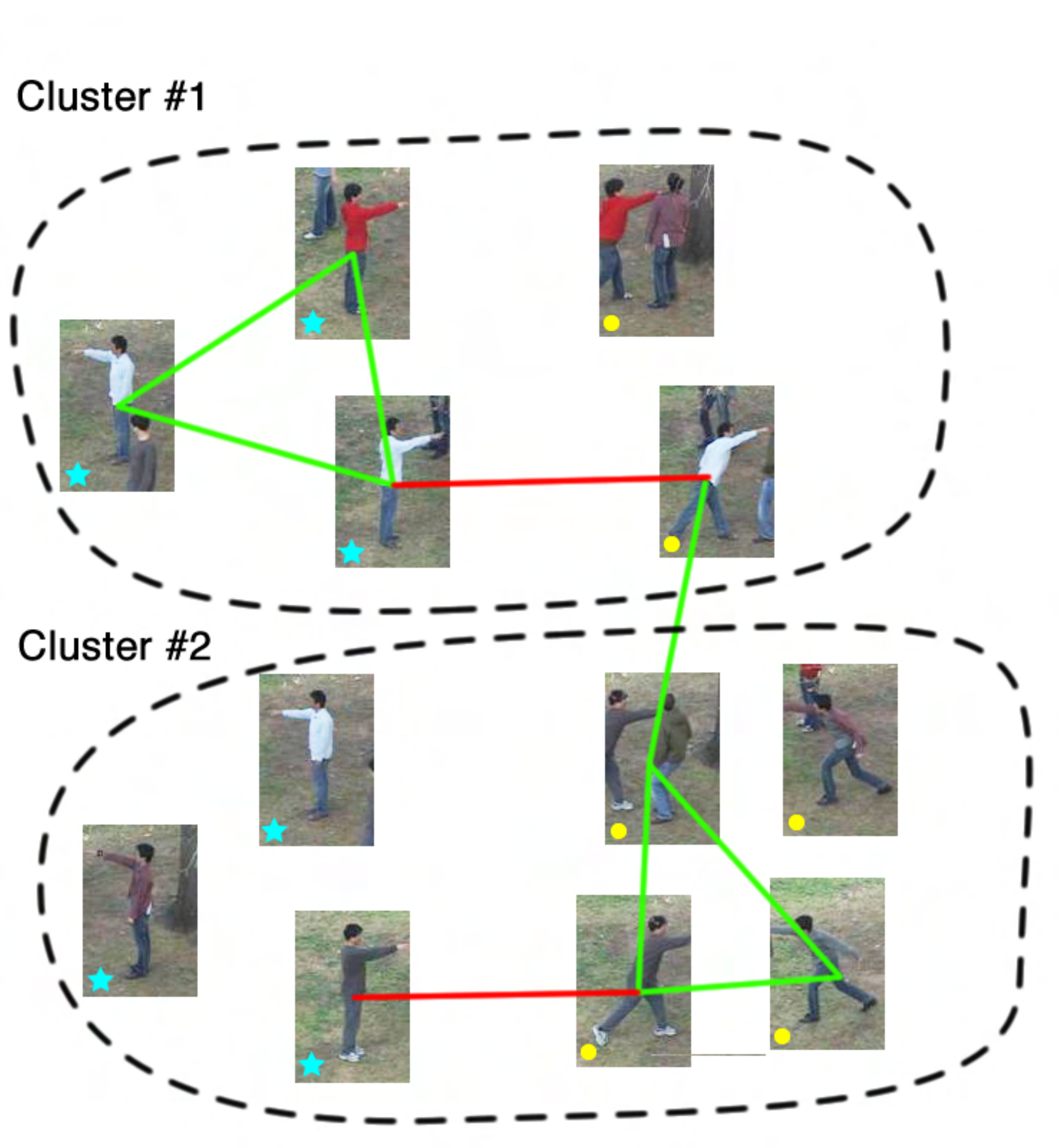}\label{fig:iter0_feedback}}
\quad
\subfloat[][First iteration with user provided constraints]
{\includegraphics[scale=0.2]{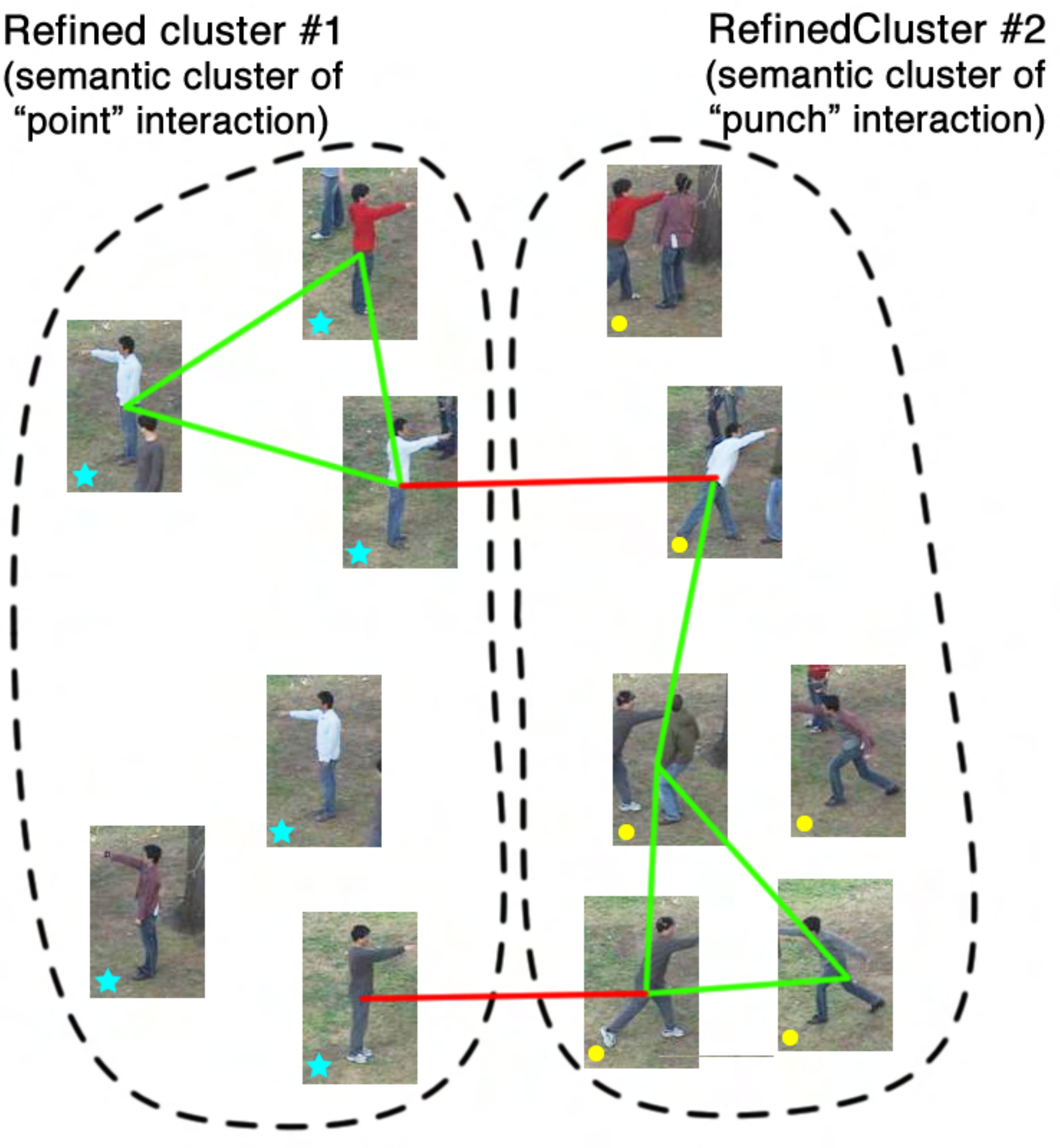}\label{fig:iter1}}

\caption{Overview of our iterative algorithm. \protect\subref{fig:iter0} 
Unsupervised step: Data are clustered into groups with large margins. 
Colored shape on the bottom left corner of each sample image indicates 
the ground truth label, which is unknown for the algorithm. 
\protect\subref{fig:iter0_feedback} User feedback: The user specifies the 
\textit{must-link} (green lines) and \textit{cannot-link} (red lines) constraints. 
\protect\subref{fig:iter1} Cluster refinement: Clusters consistent with 
\textit{must-link} and \textit{cannot-link} constraints are regenerated. }
\label{fig:iterations}
\end{figure*}

\subsection{Unsupervised Activity Recognition}


A diverse set of unsupervised methods has been developed for activity analysis, ranging from pixel-level flow models to
the clustering of person trajectories. In general, holistic scene models are
deemed to have the advantage of being more robust compared to
tracking-based approaches, because of the challenges in tracking
individual people. But, they are typically limited in the level of
semantic detail that can be modeled.  Examples of work in this area
include Zhong et al.~\cite{ZhongSV04}, who performed novelty detection
in a clustering framework based on long videos represented using
spatio-temporal derivatives.  Mehran et al.~\cite{MehranOS09}
developed a social force model for interpreting the behaviour of
crowds of people observed from a long distance.

Other related methods typically model small patches
of scenes. Hospedales et al.~\cite{HospedalesGX09} and Wang et al.~\cite{WangMG09} build novel topic
models for the actions of people or vehicles in surveillance scenes.
Kuettel et al.~\cite{KuettelBVF10} model temporal evolution of
discovered topics or activities, for instance discovering different
phases of activity.  

More closely related to our approach are those that try to form clusters of 
human activities using unlabeled data.  Niebles et al.~\cite{NieblesWF08} use a topic model over
a bag-of-words representation from local features around a person.
Wang et al.~\cite{WangJDLM06} cluster images using shape features to
discover action classes.  Calderara et al.~\cite{CalderaraHPCT11}
track individuals and reason about scenes to find anomalous
trajectories.  In this work we develop a clustering framework that
examines trajectories in an unsupervised fashion, but reasons about
interactions between these trajectories of individuals and objects.

\subsection{Clustering Methods}

Clustering is a widely-studied problem; many standard clustering
methods have been created, such as k-means, spectral
clustering~\cite{ng2002spectral}, topic models~\cite{blei03_jmlr}, and a
variety of mixture models.
In addition to the aforementioned methods, maximum margin clustering
(MMC)~\cite{xu2004maximum} emphasizes the separation between classes.
MMC is an extension of max-margin supervised learning (i.e. SVMs).
Given a set of observations, MMC performs clustering by finding the
hyperplanes with maximum margin through the data. Experimental results
have shown that this method often outperforms competing clustering
methods.

Supervised large margin methods usually lead to convex optimization
problems, while solving unsupervised versions require untangling a
non-convex integer program. Therefore, recent research tackled the
problem of reducing the computational complexity of
MMC~\cite{zhao2008efficient,zhang2009maximum}. Zhang et al.\ directly
optimize the non-convex problem by changing the loss function to
Laplacian loss, instead of optimizing the problem as a non-convex
semidefinite program (SDP)~\cite{zhang2009maximum}. Zhao et al.\
accelerated the convergence of MMC via a series of tighter relaxed MMC
instances~\cite{zhao2008efficient}. Another line of work is
incorporating further information and constraints into MMC. Hu et
al.~\cite{hu2008maximum} added slack variables for soft pairwise
constraints.  Zhou et al.\ developed a maximum margin framework that
handles unobserved knowledge in data using latent
variables~\cite{zhoulatent}.  We build on this line of work,
developing a variant of this approach and a novel model for
unsupervised discovery of human interactions.

\section{Clustering Human Interactions}
\label{sec:cluster}
We assume that an object detection and tracking algorithm exists and a 
set of trajectories are available\footnote{In section \ref{sec:experiments}
 we provide dataset-specific details on these algorithms.}. Therefore, the goal
is to cluster human trajectories
based on their interactions with surrounding humans or vehicles.
Each cluster should contain a semantically similar set of interactions.
A common approach to this problem is to feed features extracted
on each person to a standard clustering algorithm.

However, clustering interactions using a standard approach may not necessarily
result in clusters of semantically similar interactions.  Two key
reasons are:
\begin{itemize}[leftmargin=*]
\item {\bf Feature representation}: The underlying features should represent the desired semantic
similarity. Otherwise, grouping similar interactions in the space of low-level features
cannot guarantee the formation of coherent high-level clusters.
\item {\bf Lack of supervision}: A purely unsupervised clustering
  algorithm is still prone to mistakes due to intra-class
  variation in high-level semantic classes.
\end{itemize}


The proposed algorithm can handle those issues effectively. An overview of the
algorithm is presented in Fig.~\ref{fig:iterations}.  We show that by injecting a small amount of
user-provided feedback, errors in unsupervised learning can be
corrected. Leveraging latent variable representations can address the
feature representation issues. We formulate those ideas in a
novel variant of max-margin clustering.


\subsection{Max-Margin Clustering with User Feedback}

We propose a novel iterative clustering approach that improves the
quality of clusters by iterations of obtaining user feedback on
automatically-generated clusters. The basic idea is that a small
amount of feedback in each iteration not only fixes mistakes in the
clusters, but also can be generalized to other incorrectly clustered
examples.  This feedback will reduce mistakes in clustering, cases where
interactions are assigned to clusters whose dominant interaction type
is semantically different (c.f.\ cluster {\em purity} measurements).

Assume that we have a set of clusters formed from a video
dataset (Fig.~\ref{fig:iterations}(a)).  A user can be asked to view
the generated clusters and to mark a few examples, such as those
corresponding to the dominant interaction in each cluster, or
misplaced examples (Fig.~\ref{fig:iterations}(b)).

Some user-marked samples represent correctly clustered
interactions that are semantically similar.  Thus, in further
clustering they must be grouped together.  We represent these
interactions in each cluster as {\em must-link} constraints.

Interactions that are in incorrect clusters can be moved by a user to
their corresponding correct clusters.  This implies that these samples
and the ones in the must-link groups of the incorrect cluster should
never be grouped together. This can be represented as {\em
  cannot-link} constraints formed between every pair of incorrectly
clustered samples and samples in the must-link groups.  Second, a
must-link constraint should be formed with the samples in the correct
group.

In summary, the user-provided feedback indicates a few samples that
are correctly clustered and a few samples that should be moved to
another cluster in order to improve the clustering quality. This
feedback is collected iteratively and the clusters are re-generated
(Fig.~\ref{fig:iterations}(c)), resulting in pure clusters after a few
iterations.

\subsubsection{Formulation}
We modify the recently proposed
latent max-margin clustering (MMC)~\cite{zhoulatent} to formulate our clustering idea.  
MMC extends the principle of maximum margin in supervised learning (e.g. SVM) to
unsupervised clustering, where the labels of data are
unobserved. Given a set of examples $X=\{x_1, x_2, \dots, x_N\}$, the
goal of the algorithm is to find a set of binary labels $\mathcal{Y} =
\{y_{it}\} (i\in \{1, \dots, N\}, t \in \{1, \dots, K\})$. MMC groups
the data into $K$ clusters in such a way that the margin between
classes is maximal. This formulation is extended to include latent
variables which can modulate the feature representation for each data
sample.  In this case, the features for each example are altered by the
notion of latent variables such that the separation between clusters
is maximized. However, neither MMC nor latent MMC is capable of
incorporating user feedback while discovering clusters of
similar interactions. Here, we propose a novel extension of the latent
max-margin clustering framework that is able to collect feedback from
a user on a set of clusters in order to improve their quality
iteratively.

The must-link and cannot-link constraints respectively indicate a set of points
that must and must not be grouped together. 
The set of all must-link constraints is represented using $G=\{g_m\}_{m=1}^{M}$ where
$g_m \subset \{1,2,..., N\}$ indicates the indices of samples that must
be assigned to the same cluster as indicated by user.
Similarly, the cannot-link constraints
are represented using a set of pairs $C = \{(p, q)\}$
where $(p, q)$ for $p,q \in \{1,2,..., N\}$ indicate
indices of examples that must not be assigned to the same cluster.
In addition to the cluster labels $\mathcal{Y}=\{y_{it}\}$, a set of new binary variables
$\mathcal{E}=\{e_{mt}\}$ for each group and cluster is defined.

Our proposed clustering framework is defined as the
optimization:
\footnotesize

\begin{eqnarray}
&\hspace{-1.0cm}\underset{\mathcal{W,Y,E,\xi} \ge 0}{\min} &\frac{\lambda}{2}\sum_{t=1}^K ||w_t||^2 + \frac{1}{K}\sum_{i=1}^{N}\sum_{r=1}^{K}\xi_{ir}\label{eq:1}\\
&\hspace{-1.0cm} s.t.&    \sum_{t=1}^{K} y_{it} f(x_i; w_t) - f(x_i;w_r) \ge 1
- y_{ir}-\xi_{ir}  \quad \forall i,r \label{eq:2} \\
&& \sum_{t=1}^{K}y_{it} = 1 \quad \forall i \; ,  \qquad  \sum_{t=1}^{K}{e_{mt}} = 1  \quad \forall m \label{eq:3}\\
&& y_{it}\in \{0,1\} \quad \forall i,t \qquad , \quad  e_{mt}\in \{0,1\} \quad \forall m,t \label{eq:5} \\
&&  L \le \sum_{i=1}^{N} y_{it} \le U  \quad \forall t \label{eq:4} \\
&& y_{it}=e_{mt}  \quad \forall m, i \in g_m \label{eq:must_link} \\
&& y_{pt} + y_{qt} \leq 1 \quad \forall (p, q), t \label{eq:cannot_link}
\end{eqnarray}

\normalsize

\textbf{Objective Function:} 
In this formulation $\mathcal{W}=\{w_t\}_{t=1}^K$ contains
the parameters of the model.
The slack variables $\xi = \{\xi_{ir}\},
i \in \{1,\dots,N\}, t \in \{1, \dots, K\}$ allow a soft margin, and constant $\lambda$ 
controls the trade-off between the slack variables and the margin. The objective function 
(Eq.~\ref{eq:1}) and the constraint in Eq.~\ref{eq:2} optimizes the parameters of the clustering model
$f(x_i; w_t)$, and the cluster assignment variables $\mathcal{Y}$ and
$\mathcal{E}$ such that the margin between the score of the assigned
cluster for each sample and its score for any other cluster is maximum.
Here, $f(x_i; w_t) = \max_h \left[ w_t^{\top} \phi(x_i,h)\right]$
represents the score of assigning the example $x_i$ to the
cluster $t$, which is computed using the best configuration
of latent variables.
The feature vector for example $x_i$  with a latent variable configuration $h$ is denoted
by $\phi(x_i, h)$.  
$y_{it} = 1$ denotes that the example $x_i$ belongs to the
cluster $t$, $y_{it}=0$ otherwise.
Similarly $e_{mt} = 1$ denotes that the must link group $g_m$ belongs to the
cluster $t$, $e_{mt}=0$ otherwise.

\textbf{Assignment Constraints:} The constraints in Eqs.~\ref{eq:3}
and \ref{eq:5} enforce the instances (or a whole must-link group) to
necessarily be assigned to a cluster and only one cluster.

\textbf{Cluster Balance:}
The constraint in  Eq.~\ref{eq:4} avoids 
a degenerate solution to the optimization problem, where all the data
points are grouped into one cluster that has infinite margin with
other clusters. This constraint sets upper ($U$) and
lower ($L$) bounds on the size of the clusters and can further enforce
balanced clusters. 

\textbf{Must-Link Constraints:}
The constraint in Eq.~\ref{eq:must_link} ensures that all instances
in a must-link group have the same cluster label. Note that here
the same must-link group assignment variable $e_{mt}$ is shared
between all instances of a group.

\textbf{Cannot-Link Constraints:}
The constraint in Eq.~\ref{eq:cannot_link} enforces that two
cannot-link instances are not assigned to the same cluster.
Assuming $(p,q)$ represents two cannot-link samples, if they were assigned to the same cluster, we
would have $y_{pt} + y_{qt} = 2$ for at least one cluster.

\subsubsection{Optimization}
We use an alternating descent algorithm
to solve the optimization problem in Eq.~\ref{eq:1}
considering the constraints defined in Eqs.~\ref{eq:2}-\ref{eq:cannot_link}.
This minimization involves solving for unknown latent
variables $h$ and cluster assignments $y_{it}$, and then revising
estimates of parameters $w_t$.  We use
the non-convex regularized bundle method
(NRBM)~\cite{do2009large}.  Details of the initialization strategies
are described in the experimental results.

We can obtain the set of must-link and cannot-link constraints
iteratively from a user.  In the first iteration, a clustering of
interactions is generated with no supervision, i.e.\ without considering
any constraint of this type. The initial clustering is presented to
a user to obtain his/her feedback. The feedback is modeled as additional
constraints, as described above. Then, the samples are clustered
again in the next iteration to generate new groups of human interactions
that reflect the cumulative user-provided feedback in all previous iterations. By
iteratively clustering and obtaining feedback one can construct a pure
clustering of data with no incorrectly clustered samples.  In the
experiments section we will show that this can be achieved with a small
amount of user feedback.

\section{Features and Implementation Details}
\label{sec:features}

We develop methods for clustering human actions according to their
interactions.  The framework outlined in Sec.~\ref{sec:cluster} is a
general-purpose approach that could be used in a variety of settings
for analyzing human interactions.  For concreteness, we evaluate our
algorithm for human interaction clustering on three standard datasets
-- UT-Interaction~\cite{RyooA09}, Collective
Activity~\cite{choi2011learning}, and VIRAT~\cite{oh2011large}.
UT-Interaction and Collective Activity
are standard datasets, providing
well-defined sets of activity classes for measuring clustering performance.
VIRAT contains a larger, more diverse set of
potential interactions between humans or between humans and vehicles.
It provides an excellent domain on which to evaluate algorithms'
abilities to discover classes of interactions that are not defined a
priori.



We utilize feature representations appropriate to each dataset.  For
the Collective Activity Dataset, we analyze the human detections in a
frame, and cluster video frames according to the group activity
present.  We describe each frame using an existing method that
represents the appearance of person in a scene using HOG
features~\cite{LanWYRM12}.  These HOG features are classified into
categories of pose/action, the values of which are treated as latent
variables in the clustering model.

For clustering human trajectories in the VIRAT and UT-Interaction
datasets, we develop a set of features including relative
position/velocity and appearance.  These are augmented with a latent
variable representation that handles temporal alignment.  Details of
these features are provided next.

\subsection{Proximity Features}

Given a set of trajectories of people in a scene, we wish to build a
representation for their interactions.  We assume we have trajectories
for the people and objects of interest (e.g. vehicles) in a
scene.  Different classes of interaction will likely have
stereotypical patterns of proximity.  For instance, a crowd of people
might stand together, engaged in a conversation.  Two people
might walk together across a scene.  A solitary person might approach
a parked vehicle.  We build a representation that captures the
relative positions and velocities of people in a scene in
order to differentiate between these types of categories of
interaction.

We build a representation for each person in a scene.  Focusing on one
person, we examine his positions and movements with respect to other
people and vehicles in the scene.  We use a
representation that only examines the focal person and the one person
and one vehicle that is closest to that focal person over the course
of a trajectory.  For that one person or vehicle, we build a histogram
representation that captures the relative position and velocity of the
focal person with respect to the other.

The histogram representation requires choosing a quantization with
respect to relative velocity and distance.  In order to reduce
dependence on an a priori specification of these bin edges, we use an
unsupervised approach.  We collect sets of samples of relative
velocities and distances across a dataset, and then build either a
mixture of Gaussians model or a percentile-based representation in
order to construct the histogram representation.  Each sample point
from a respective trajectory of person or vehicle is encoded according
to its responsibility under each component of the mixture of Gaussians
or its membership in a percentile range.

More precisely, for a person trajectory $x$, the magnitude of its velocity,
is estimated via finite differences between the start and end locations
of the track. Then a histogram of velocity is created using soft quantized
Gaussian Mixture Model. Similarly, the relative distance between person $x$ and its
nearest person trajectory at time $t$ is hard quantized to set percentile-based
bins, and the histogram of distance is computed by summing
over all times.

\subsection{Appearance Features}

Beyond relative positions and movements, the appearance of a person
can capture information about the type of interaction occurring.  We
augment the track-level proximity features with appearance features
based on histogram of oriented gradients (HOG) and histogram of
oriented flow (HOF) features.
In the UT-Interaction dataset, we use a mixture of Gaussians model
to represent appearance.  We concatenate the HOG and HOF features into
a single feature vector and then train a mixture of Gaussians model.
Again, each frame is represented by its responsibilities under this
model, and summed over time to create a representation for the
trajectory.

\subsection{Latent Variables for Temporal Alignment}

The aforementioned features describe a trajectory via a combination of
distance, velocity, and appearance features.  However, a challenge
when attempting to cluster person trajectories is alignment between
different tracks.  Global histogram-type features of this type can be
used to represent trajectories.  Yet this type of representation will
suffer from a lack of alignment between features for different tracks.
For instance, a person might spend a portion
of a trajectory standing still, before engaging in an interaction.
The precise start or end points of this period of
motion are variable, and can be modeled with a latent variable.

In order to account for these differences, we modulate the track
features defined above with latent variables that can be used to align
the features of different trajectories.  We include latent variables
to offset the temporal range on which relative distance and velocity
features are defined.  

\section{Experiments}
\label{sec:experiments}

\textbf{Performance measure}: We measure clustering performance
using {\em purity}, a standard measure which evaluates accuracy of
most frequent class in each cluster. In each cluster if we assume the
points that have the same label as the most frequent class are
\textit{correctly} labeled, then the purity is the ratio of all
correctly labeled points to the total number of points. Note that
purity is analogous to classification accuracy in a setting where the
number of clusters equals the number of ground truth classes.

\textbf{Initialization:} For the first iteration, which is fully
unsupervised, we initialize our clustering algorithm with a weight
vector with all weights set to 1. This produces a set of clusters,
then we obtain feedback from the user and add the constraints to our
clustering algorithm. For the next iteration, we initialize the
algorithm with the weight vector that we obtained from the previous
iteration. We do this iteratively until we reach 100\% purity.



\subsection{Datasets}

{\bf UT-Interaction Dataset:} The UT-Interaction
Dataset~\cite{RyooA09} contains 2 sets of videos containing pairs of
people interacting with each other.  Set 1 is captured in a parking
lot with a stationary background.  Set 2 is captured on a grassy lawn
with slight background movement and some camera jitter.  Each set
contains 10 video sequences with at least one occurrence of each of 6
categories of interaction: {\em shake hands}, {\em hug}, {\em kick},
{\em point}, {\em punch}, and {\em push}.  We use the classification
version of the dataset, and run automated human
detection~\cite{DalalT05} and tracking~\cite{RossLLY08} to obtain
trajectories of the two people involved in each interaction.  Set 1
exhibits scale variation, and the scale of the humans in each sequence
is automatically estimated from human detection results.  We compute
velocity, distance, and appearance features for each person
(Sec.~\ref{sec:features}).

We use two different latent variables in our 
experiments on UT-Interaction. 
The first is a temporal alignment latent variable that chooses
the best 20 frame long temporal window from a track.
The second latent variable models who is playing which role in an
interaction -- for example in a pushing interaction, one person is the
pusher, and the other the ``pushee.''  A latent variable is used to
swap the roles of the two people in the feature vector.
Since the UT-Interaction dataset is cleanly structured, with each
interaction coming from one of 6 categories, we cluster the tracks
into 6 groups. We conduct experiments using a variety of values for 
parameter $\lambda$ in the set of 
$\{10^{-3}, 10^{-2}, ..., 10^2, 10^3\}$, and the best purity is 
selected. We set lower bound ($L$) and upper bound ($U$) of clusters to 
$0.9$ and $1.1$ of \textit{average cluster size} respectively.

\textbf{Collective Activity:}
This dataset contains a total of 44 short video clips recorded by 
consumer camcorders. In each video, people are annotated 
every ten frames, and labeled as one of the following five categories: 
\textit{crossing, waiting, queuing, walking,} and \textit{talking}. The 
label of each frame is assigned according to the dominant activity of 
people in that frame. The features are obtained from \cite{LanWYRM12}:
from each activity category one third of the 
videos are taken to be clustered using our model, and the rest used
to for the joint action/pose classifiers that are used as features.

Each person can have one of the following eight pose categories: 
\textit{right, front-right, front, front-left, left, back-left, back}
and\textit{back-right}. We assign an action label to each person 
according to his/her pose and activity. Therefore, there are forty 
different action labels (e.g. \textit{crossing front-left}). These action 
labels are latent variables and our algorithm automatically assigns them 
to people. We cluster the scenes into $K=6$ clusters. In our experiment
we tried a wide range of values for $\lambda$ in the set of
$\{10^{-3}, 10^{-2}, ..., 10^2, 10^3\}$ for both the first iteration of
our algorithm and MMC. We used the best purity for comparison. 
Lower bound ($L$) and upper bound ($U$) are set to $0.6$ and $1.4$
of \textit{average cluster size}, respectively.

\textbf{VIRAT Dataset:}
The release 2 of VIRAT Ground dataset~\cite{oh2011large} contains more than 8 hours of
videos captured by surveillance cameras from 12 different
scenes. 
The ground truth annotation contains rare human-vehicle
interactions designed for detection tasks in surveillance settings.
However, in this work we are interested in discovering other types of
interactions such as human-human interactions in addition to
human-vehicle interactions.  Therefore, we defined a new set of labels
and manually labeled a portion of the dataset. The label set
contains: \textit{talking to a person, interacting with a car, walking
  alone, walking with a person,} and \textit{standing alone}

In this dataset, we focus on scene 0001, viewing a parking lot. We used a state-of-the-art tracking algorithm~\cite{zhang2014fast} to automatically extract 90 human/vehicle 
tracklets of length 12 seconds (80 humans and 10 vehicles) from manual initializations.
We formed the ground truth by labeling the human tracklets based on their 
interaction with other people or vehicles. 

We compute distance and velocity features over each quarter of each tracklet, i.e.\ temporally
binning features with 4 replicates.  The dataset contains scenes with
multiple people and vehicles present at once.  For each focal person,
we find the one person and one vehicle with shortest median distance
to the focal person,
 which are considered as the closest person 
and vehicle to the focal person, respectively. 
Distance features are computed with respect
to this vehicle and person pair.  Latent variables for temporal
alignment are used in a sliding window fashion, choosing a 6 second 
long sub-region within the tracklet. We set the number of clusters $K=5$, 
lower bound $L=0.4$, and upper bound $U=1.6$.
We use the best purity $\lambda$ in the set $\{10^{-3}, 10^{-2}, ..., 10^{2}, 10^3\}$ for
each method.

\begin{figure*}
\centering
\subfloat[][UT set 1]
{\includegraphics[scale=0.19]{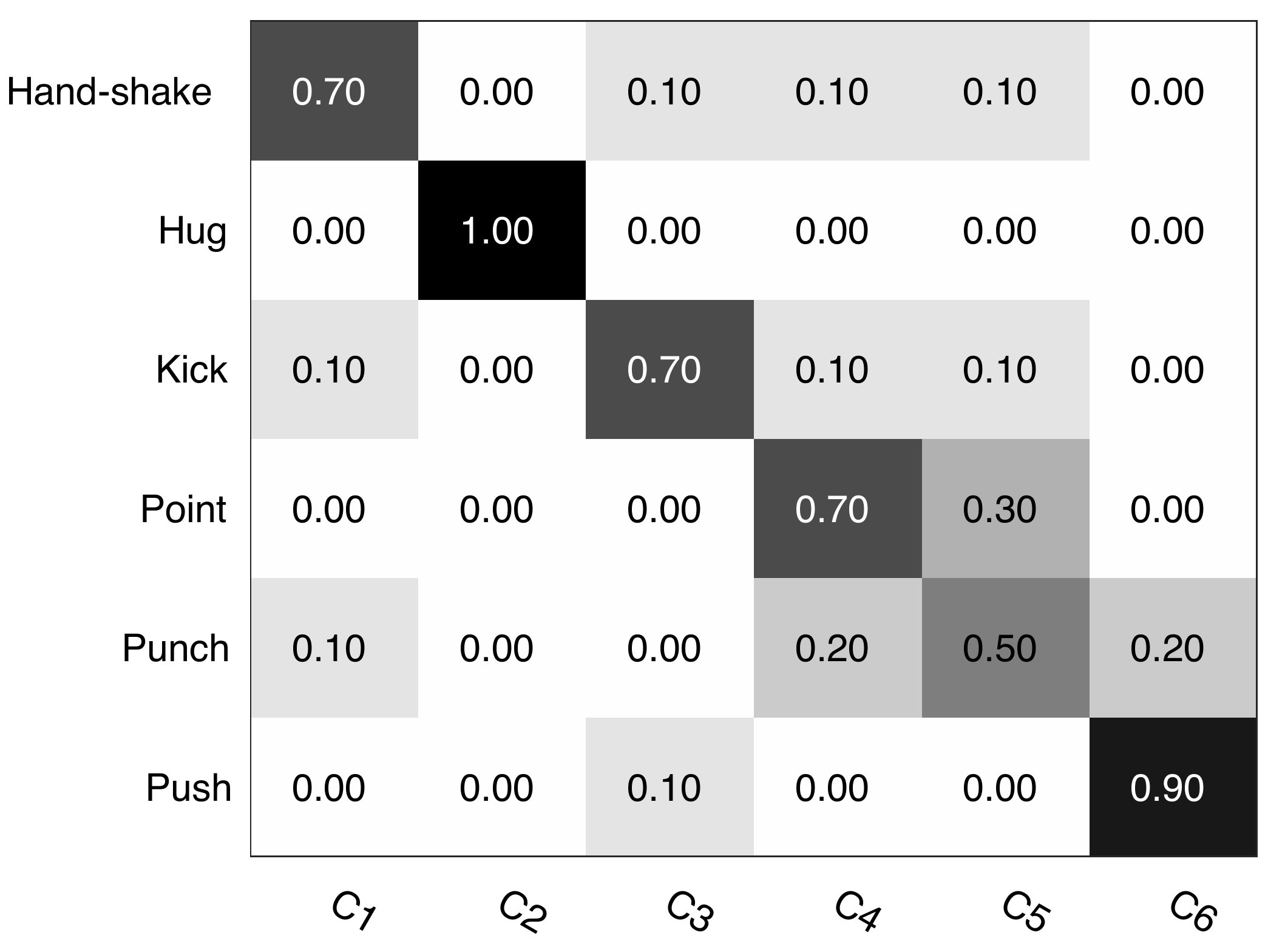}}
\label{fig:ut_set1_conf_mat}
\subfloat[][UT set 2]
{\includegraphics[scale=0.19]{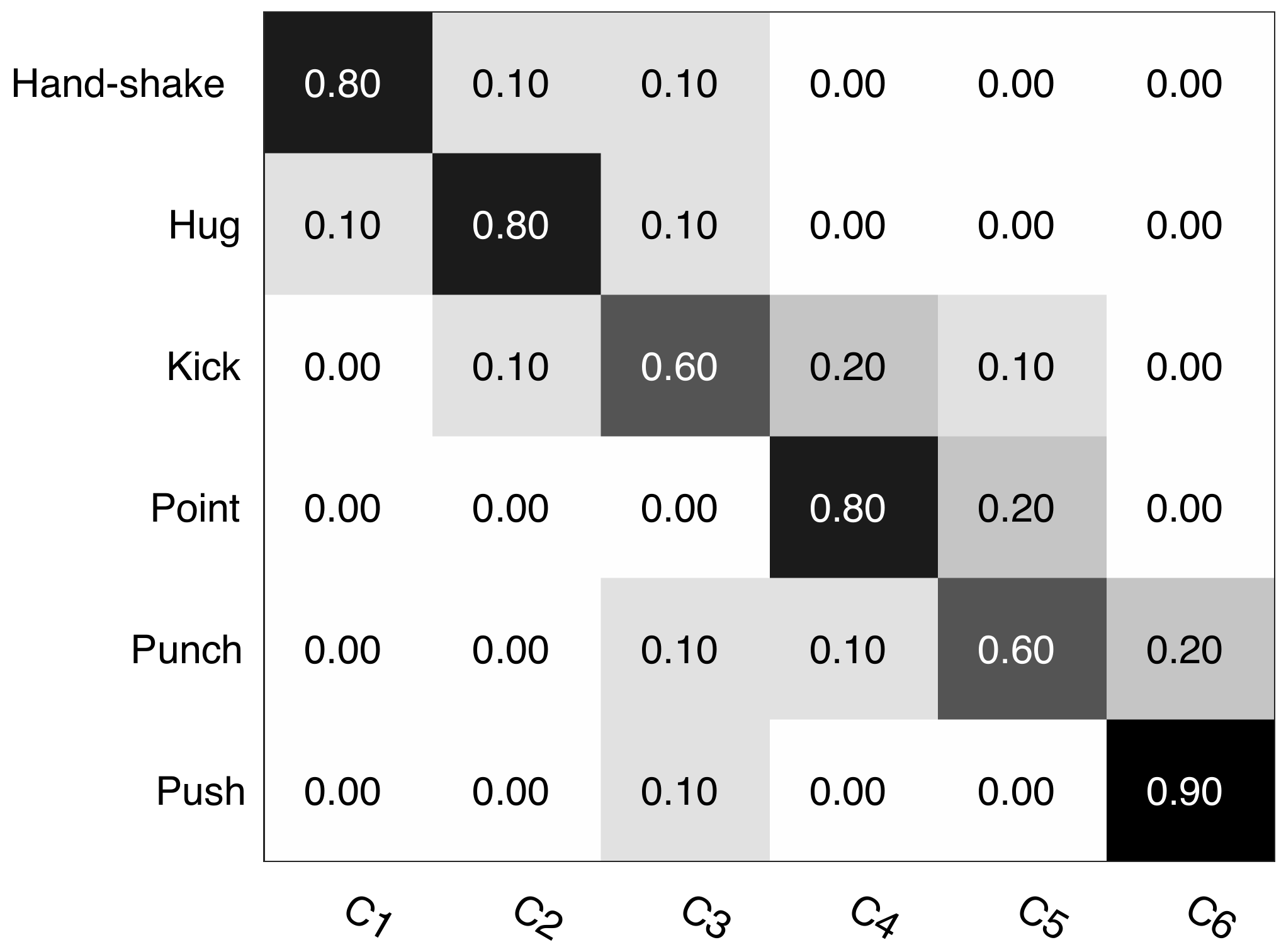}}
\label{fig:ut_set2_conf_mat}
\subfloat[][Collective Activity]
{\includegraphics[scale=0.28]{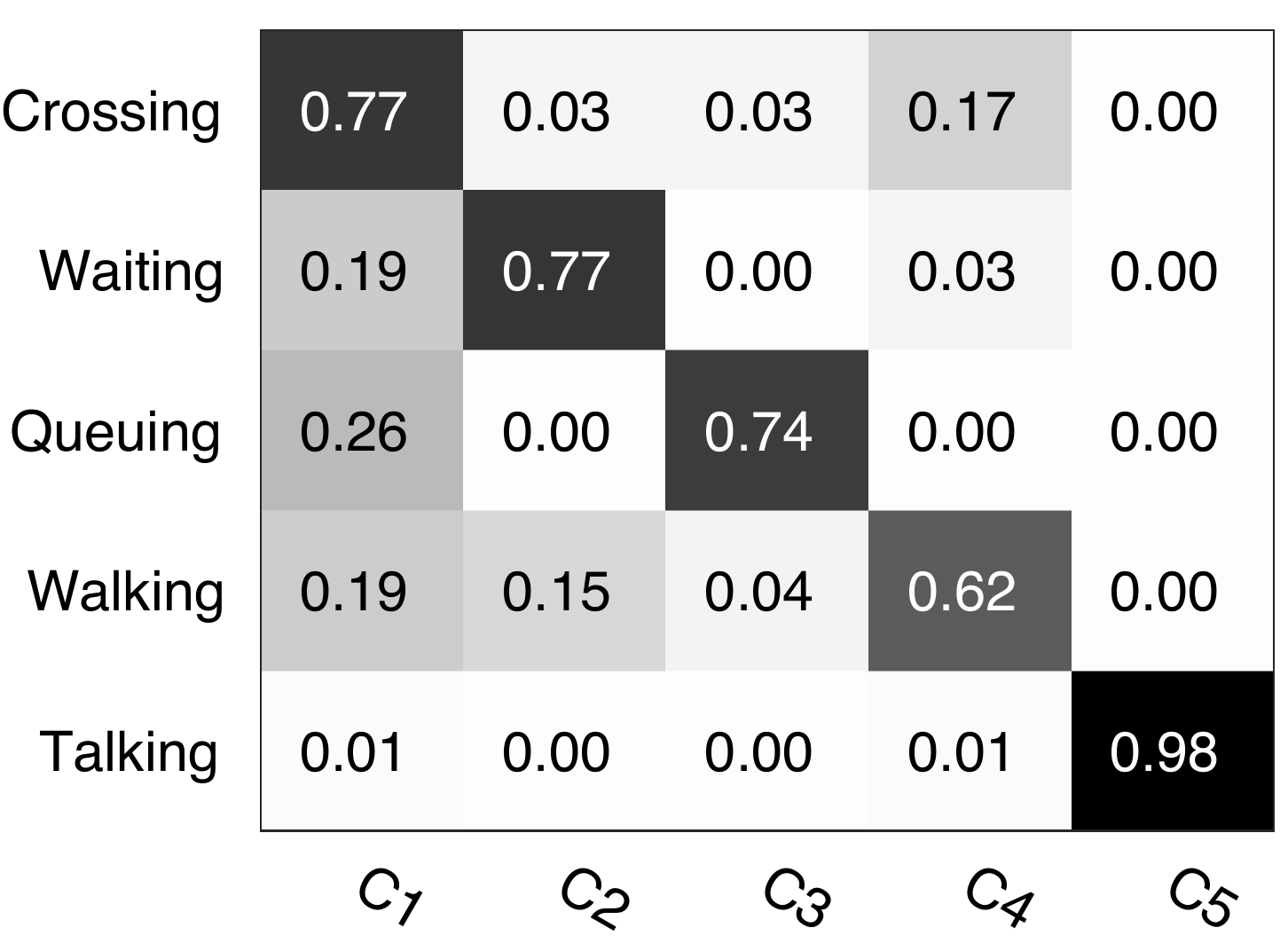}}
\label{fig:ca_conf_mat}
\subfloat[][VIRAT]
{\includegraphics[scale=0.28]{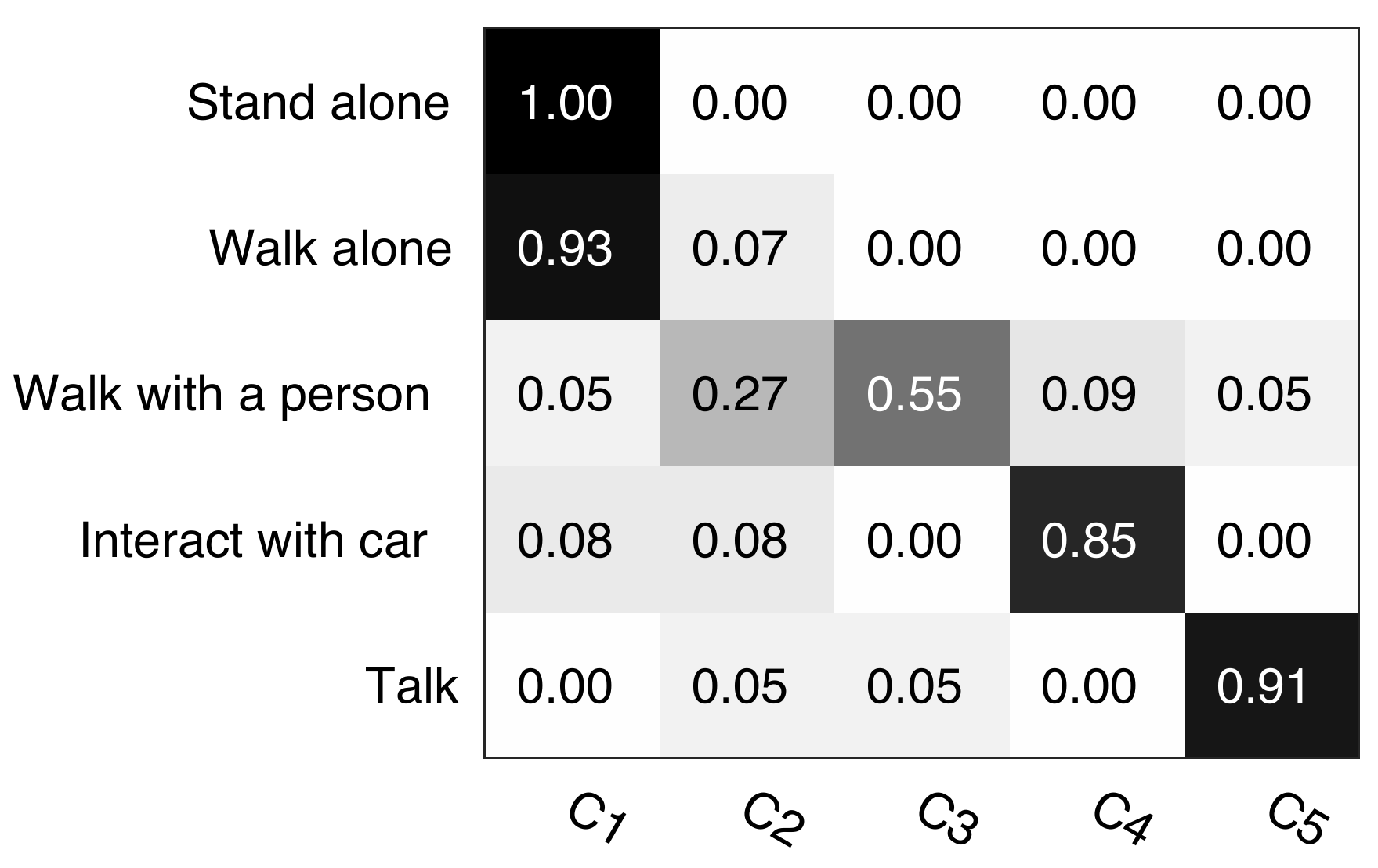}}
\label{fig:virat_conf_mat}

\caption{Confusion matrices of clusters generated at
  iteration-0 for UT Interactions Set 1 (a), Set 2 (b), Collective
  Activity dataset (c), and VIRAT(d).}
\label{fig:conf_matrices}
\end{figure*}

\begin{figure*}[]
\centering
\subfloat[][UT set 1]
{\includegraphics[scale=0.29]{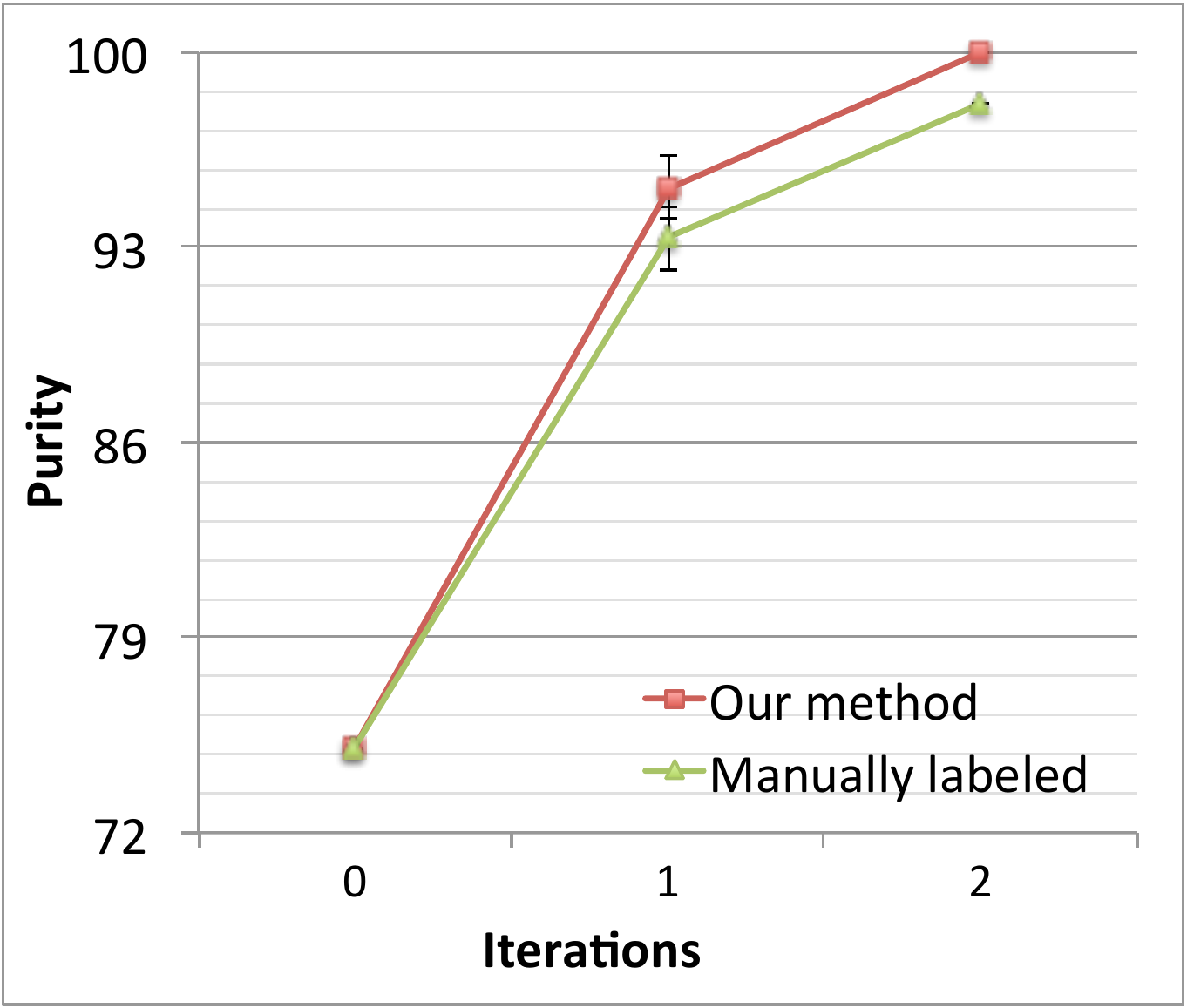}\label{fig:ut_curve_set1}}
\quad
\subfloat[][UT set 2]
{\includegraphics[scale=0.29]{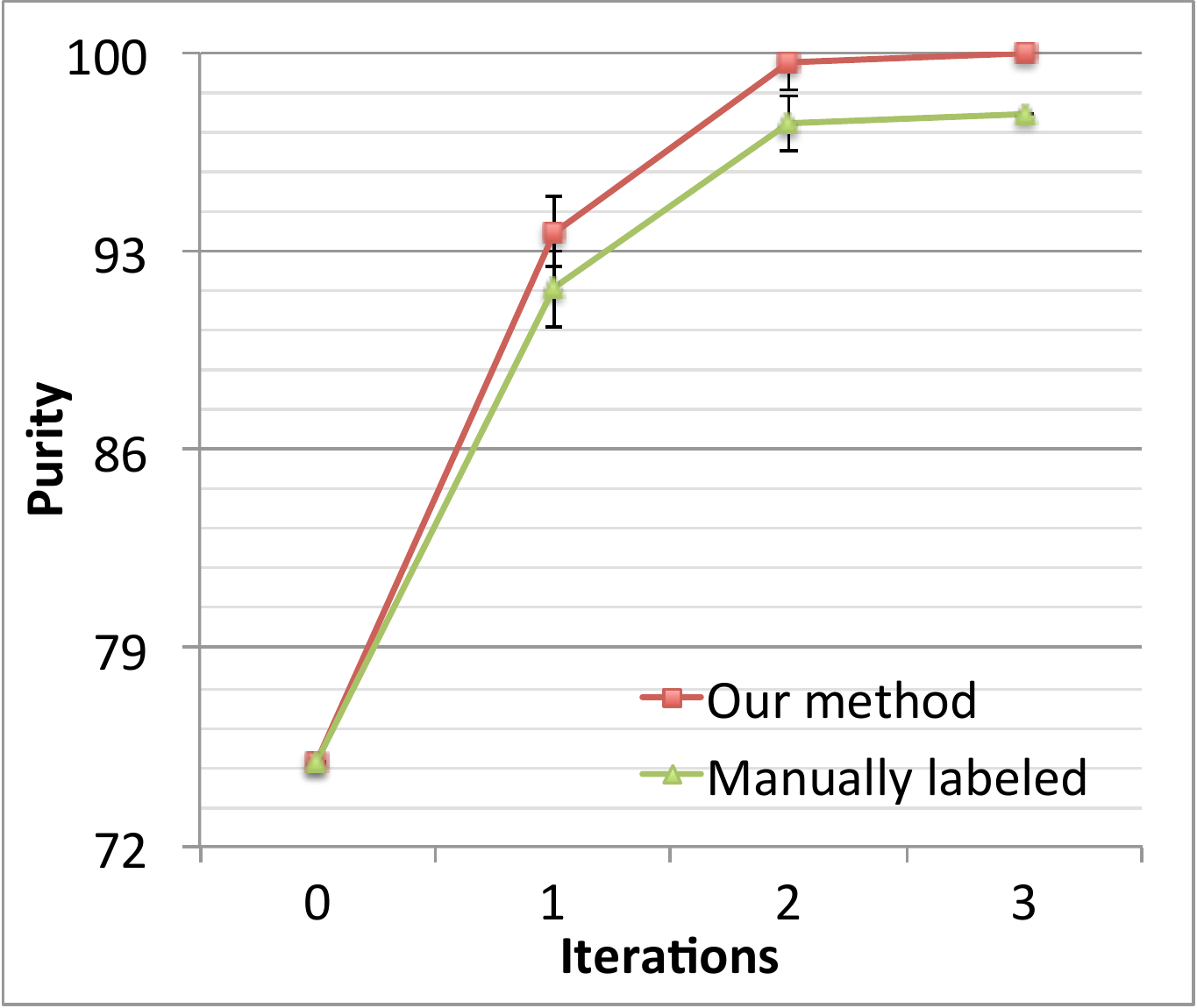}\label{fig:ut_curve_set2}}
\quad
\subfloat[][Collective Activity]
{\includegraphics[scale=0.29]{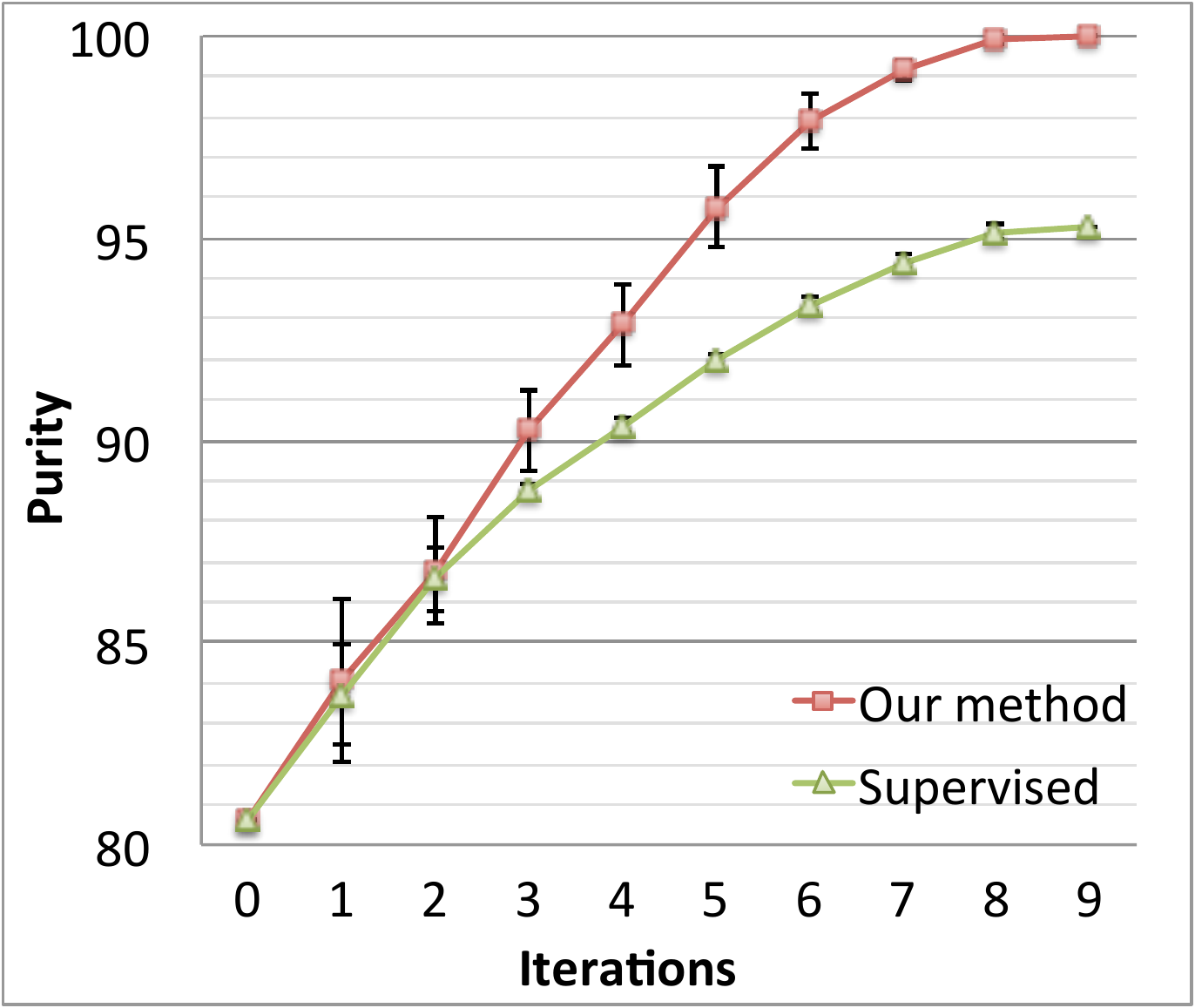}\label{fig:ca_curve}}
\quad
\subfloat[][VIRAT]
{\includegraphics[scale=0.29]{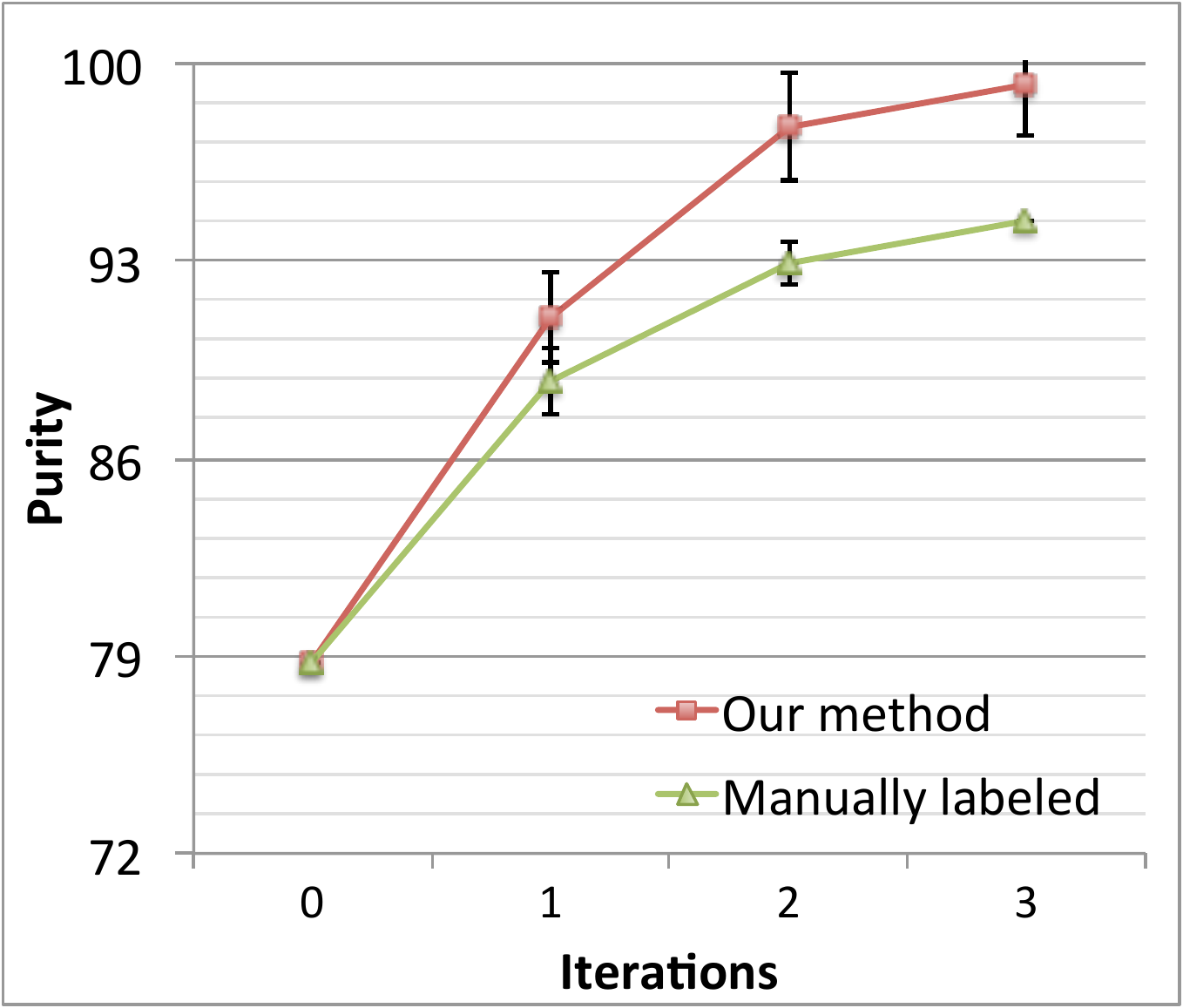}\label{fig:virat_curve}}

\caption{Average purity of the proposed clustering model
for UT-Interaction Set 1 (a), Set 2 (b), Collective Activity (c), and VIRAT (d).
Our method constructs 100\% pure clusters
after a few iterations of obtaining user-feedback.
Our performance is significantly better than
the \textit{manually labeled} baseline that uses the feedback
for correcting misclustered interactions that are
generated in iteration-0 with zero supervision. }
\label{fig:curves}
\end{figure*}

\subsection{Results}

\begin{figure}
\centering
{\includegraphics[scale=0.295]{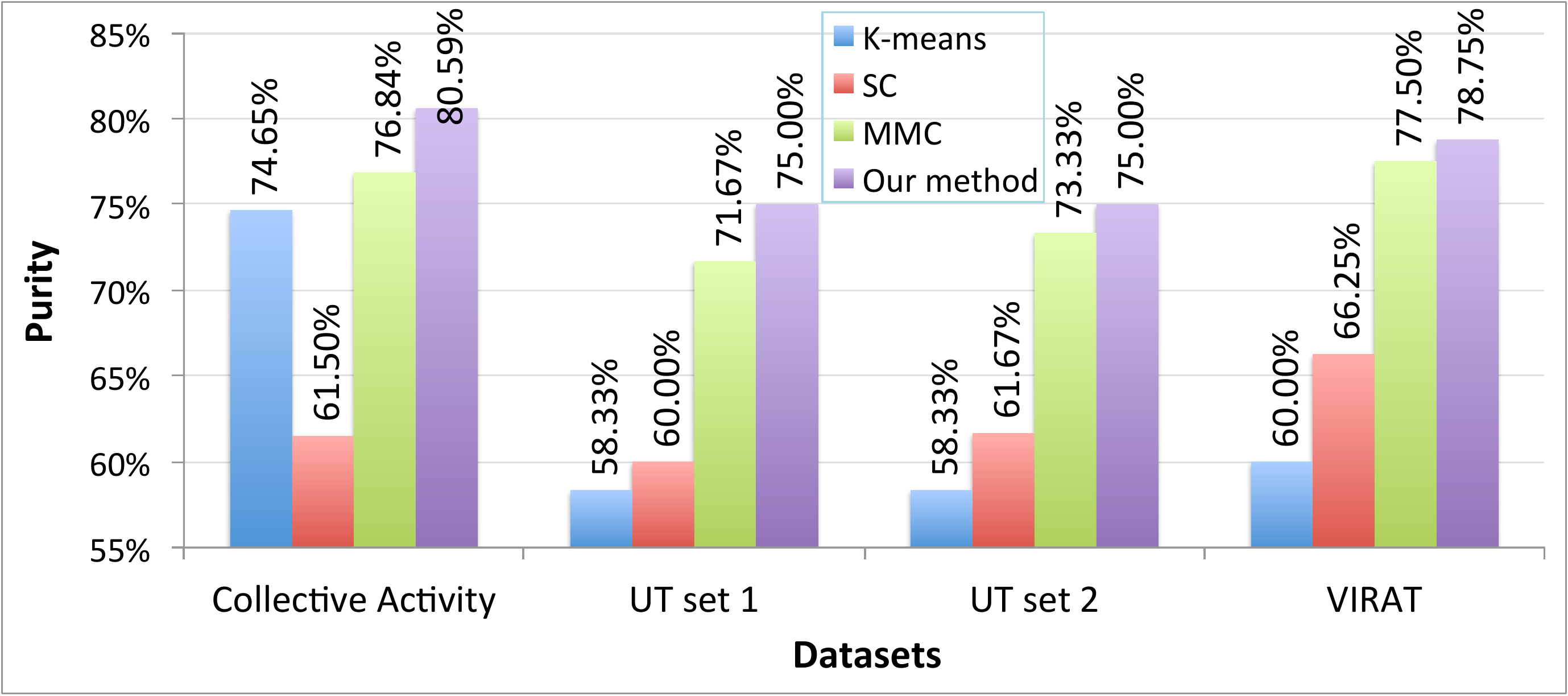}}
\caption{Performance of our clustering method compared
to other baseline methods on different datasets. For all datasets, our proposed method
generates clusters of high purity without any supervision on activity labels.}
\label{fig:unsupervised}
\end{figure}

\textbf{Fully unsupervised (iteration-$0$):} In the first round of the
process, we leverage the features and latent model to cluster data
into groups with large margin.  There is no supervision in this
step. Our experiments show that the clustering results of this step
are better than baseline clustering methods. We compare our method
with \textit{K-means, Spectral Clustering,} and \textit{Max-Margin
  Clustering}.  Fig.~\ref{fig:unsupervised} shows the results in
terms of purity, and Fig.~\ref{fig:conf_matrices} shows confusion matrices. Our method works significantly better than the common
baselines. For instance, on the Collective Activity dataset, among
baselines MMC achieved highest purity, $76.84\%$. While our method
produces clusters with $80.59\%$ purity.

\textbf{User feedback:}
After the first iteration, a user is asked to look through the clusters
and choose a small group of dominant interactions from  each cluster. 
These form the \textit{must-link} constraints.  Then we ask the user to select a 
few mis-clustered interactions from each cluster and put them in their 
corresponding groups that are chosen in the previous step. These form the 
\textit{cannot-link} constraints. Note that if the user doesn't provide 
misclustered points for some clusters, we consider them as pure clusters 
and don't break them in the next iteration.

We used the ground truth labels to mimic the user feedback. In each iteration,
$m$ interactions are selected uniformly at random from 
the dominant interactions of each cluster. Those form the \textit{must-link} constraints. We chose $m=5$ for Collective Activity and UT and $m=8$ for VIRAT, 
which can practically be done by a real user. Then, we randomly select up to $c$ interactions from 
the misclustered interactions, form \textit{cannot-link} constraints,
then add them to their corresponding groups. We set $c=5$ for
Collective Activity and $c=2$ for UT and VIRAT. 

Correcting the label of misclustered 
interactions will increase {\em purity}, since the total number of 
correctly clustered points will be increased.
In order to demonstrate how our method is capable of generalizing
the user-feedback to incorrectly labeled interactions, a baseline
method called \textit{Manually labeled} is also defined that represents
the purity of clusters after correcting the misclustered interactions
solely based on the feedback.

Fig.~\ref{fig:curves} shows the average performance of our method over
10 runs with different random samplings at each iteration.  The
results show that our proposed method generates pure clusters after a
few iterations of obtaining user-feedback on clusters that were
originally generated with zero supervision (i.e.\ iteration-0). The
comparison of our method with the manually labeled baseline also
demonstrates how our method can generalize the user-feedback to
mis-clustered samples.  Error bars show the standard deviation over the 10 runs.

\textbf{Running time:} 
%
Our proposed clustering algorithm takes only a few seconds to cluster data given user feedback.
 The average clustering time per feedback iteration for each dataset is as follows:
VIRAT: $2$ seconds, Collective Activity: $7$s, and UT-Interaction:
$18$s on a Intel Core i7 CPU (@ $3.40$GHz) in a MATLAB implementation.

\textbf{User effort:} The number of data points corrected by the user in each iteration is small.
On average the number of misclustered points that are labeled by the user in each iteration is 
$9.5\pm 3.2$ for VIRAT, $9.6 \pm 6.4$ for Collective Activity, and $4.5 \pm 4.9$ for UT. 
Note that, overall, this corresponds to a small amount of labeling compared to the dataset size.
For instance, for the Collective Activity dataset the total number of these annotations 
over all iterations is only $15\%$ of the whole dataset on average.

\section{Conclusion}
\label{sec:conclusion}

We proposed a method for discovering human interactions in video
sequences based on unsupervised learning combined with user feedback.
The method operates on trajectories of people, and reasons about their
interactions with other people and/or vehicles present in a set of videos.
We use feature representations that allow the model to account for
alignment of trajectories extracted from different parts of a video
and the actions of individual people.  A novel variant of latent
max-margin clustering was developed to discover clusters in an
iterative fashion, including user feedback at each iteration.

The method shows promise for automatically discovering the types of
interactions that occur in a scene.  On the standard UT-Interaction
and Collective Activity datasets, the purely unsupervised approach obtains cluster purity that
is close to methods based on supervised classification.  On the large VIRAT corpus, a varied set of
human-human and human-vehicle interactions were discovered.  A small
number of iterations of limited user feedback results in perfectly
pure clusters of human interactions, demonstrating a promising
alternative to supervised approaches for human interaction analysis.

{\small
\bibliographystyle{ieee}
\bibliography{egbib}
}

\end{document}